\documentclass[article]{elsarticle}

\usepackage{hyperref}
\usepackage{amsmath}
\usepackage{fancyvrb}
\usepackage{bera}

\journal{Neurocomputing}

\begin{document}

\begin{frontmatter}

\title{An Efficient Method for online Detection of Polychronous Patterns in Spiking Neural Networks}

\author{Joseph Chrol-Cannon}
\author{Yaochu Jin\fnref{email}}
\author{Andr\'{e} Gr\"{u}ning}
\address{Department of Computer Science, University of Surrey, Guildford, Surrey, GU2 7XH, UK}
\fntext[email]{yaochu.jin@surrey.ac.uk}

\begin{abstract}
Polychronous neural groups are effective structures for the
recognition of precise spike-timing patterns but the detection
method is an inefficient multi-stage brute force process that
works off-line on pre-recorded simulation data.
This work presents a new model of polychronous patterns
that can capture precise sequences of spikes directly in the neural
simulation. In this scheme, each neuron is assigned a randomized 
code that is used to tag the post-synaptic neurons whenever 
a spike is transmitted. This creates a polychronous code
that preserves the order of pre-synaptic activity and can be 
registered in a hash table when the post-synaptic neuron spikes. 
A polychronous code is a sub-component of a polychronous group
that will occur, along with others, when the group is active. We
demonstrate the representational and pattern recognition ability of
polychronous codes on a direction selective visual task involving
moving bars that is typical of a computation performed by simple cells
in the cortex. The computational efficiency of the proposed algorithm
far exceeds existing polychronous group detection methods and is well suited for online detection.
\end{abstract}


\end{frontmatter}



\section{Introduction}
Spiking neurons \cite{Paugam2012} present quite a different paradigm to those of 
artificial neural networks that work directly on real valued variables 
\cite{Maass1997}. Often, investigators choose to decode the spiking activity using 
various methods \cite{Paugam2012} into real values such that they can be used with 
traditional regression and classification algorithms \cite{Maass2002}. 

Polychronization \cite{Izhikevich2006} is a spiking model of memory and computation that 
avoids this artificial decoding process. Instead, it treats each 
causally bound cascade of spiking activity as a distinct memory or 
computation. Precisely repeating spatio-temporal patterns of spikes, 
and the underlying network structures (defined by connections, axon delays 
and synaptic weights) that facilitate them, are defined as Polychronous 
Neural Groups (PNGs). In addition to avoiding arbitrary decoding of spiking 
signals, PNGs also have the advantage of linking up with a number of seminal theories 
in neuroscience. The activation of PNGs reflect the Pattern Recognition 
Theory of Mind \cite{Kurzweil2012} in which all mental content and computation is reduced 
to the combination of a large number of pattern recognizers. When we 
consider the synaptic adaptation that is required to form the neural 
structure of PNGs, Hebbian Cell Assembly \cite{Hebb1949} is descriptive of the 
organizational process of neural representation. If we also consider the 
synaptic adaptation as occurring in a competitive environment within the 
brain, Neural Darwinism (or the Theory of Neuronal Group Selection) \cite{Edelman1987,Izhikevich2004b} 
also becomes significant in describing this formative process. There are 
recent computational neuroscience works that have used PNGs to study 
cognitive representation \cite{Vertes2010,Guise2013,Guise2015thesis,Clair2013,Clair2015}, 
as well as the basis for a model of working memory 
\cite{Szatmary2010,Ioannou2014,Ioannou2015}. In general, the PNG model of spiking 
computation presents the potential for a significantly higher capacity 
\cite{Izhikevich2006,Maier2008,Cannon2012,Ioannou2012} over previous forms of spike-coding.

Why then are PNGs not more widely employed in the application and study 
of spiking neural models? We suggest two reasons. Firstly, that it is 
enticing to integrate spiking models with machine learning methods 
instead, due to the clearer mathematical underpinnings of that field, as 
well as the recent advances and generated interest \cite{Lecun2015}. Secondly, the 
algorithms currently available to detect PNGs \cite{Izhikevich2006,Martinez2009} are inefficient and 
typically cannot be run on-line, but rather on stored spiking data.

The aim of this work is to address this second limitation by exploring a 
minimal model of polychronization that can be used to detect precisely 
recurring temporal patterns of spiking activity in a highly efficient 
manner that can execute as part of the spiking simulation and therefore 
run in an on-line fashion. It is hoped that this will contribute to a 
vein of work \cite{Guise2014,Rekabdar2015d,Sun2015,Sun2016} that is currently attempting to 
facilitate the study and application of spiking networks in their own 
terms, rather than resorting to more general machine learning frameworks.

\subsection{Polychronous Neural Groups}

\begin{figure}[ht]
\hfill
\begin{center}
\includegraphics[width=4.5in]{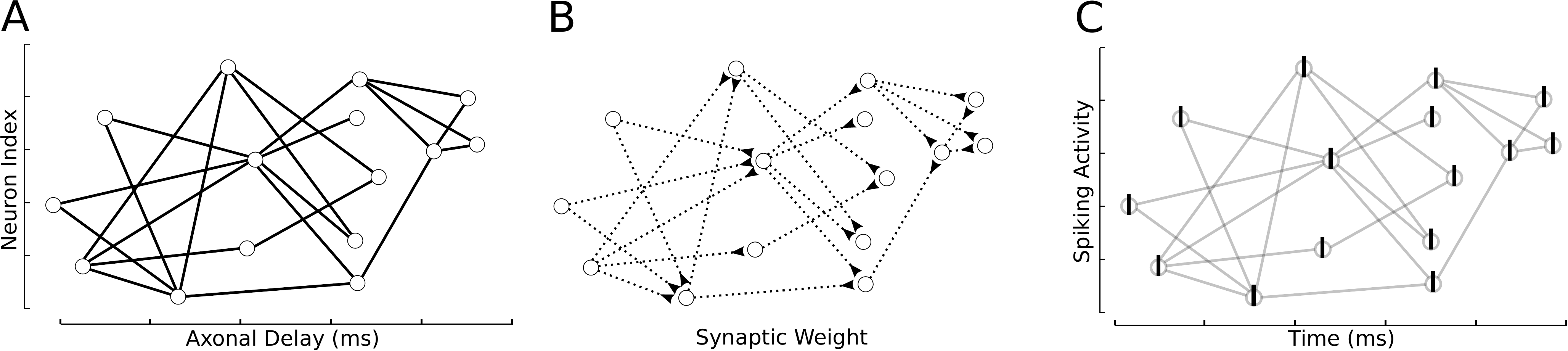}
\end{center}
\caption{Depiction of a single PNG according to its three aspects.
{\bf A:} Structural PNG, defined by its connection delays.
{\bf B:} Adapted PNG, defined by its synaptic weights.
{\bf C:} Activated PNG, defined by its set of spike-timings. }
\label{fig:png}
\end{figure}

A PNG \cite{Izhikevich2006} is a time-locked pattern of activity that cascades through a set
of neurons. Apart from the external input required to \emph{trigger} the
PNG, no further input is required to cause all constituent neurons
to fire at their precise time. The structure of a PNG is defined
at three levels, each of which are visualized in Figure \ref{fig:png}.
The potential for a PNG is determined by its structure in terms of
connectivity and conduction delays: these determine the possibility
of pre-synaptic action potentials (spikes) to arrive simultaneously and thus
cause further spikes. The adapted synaptic weights at these crucial
junctures must be strong enough to propagate enough current to activate
the PNG. Finally, the external input to the neural circuit must match
the triggering \emph{anchor} neurons in order for any PNG to become active
during simulation. Each level of a PNGs definition is dependent upon the former.

Originally, PNGs were introduced as a potential substrate for the
neural groups in the theory of Neural Darwinism \cite{Edelman1987,Izhikevich2004b,Izhikevich2006}.
From the outset, they have been demonstrated to have extremely high
computational/memory capacity both in theory \cite{Izhikevich2006} and
in practice \cite{Maier2008,Cannon2012,Ioannou2012}. Since their introduction, 
PNG's have also been applied to pattern recognition tasks in both supervised 
\cite{Paugam2006,Paugam2008} and unsupervised 
\cite{Rekabdar2014,Rekabdar2015a,Rekabdar2015b,Rekabdar2015c} forms. The 
main benefit of using a polychronous representation like a PNG is that
it retains all of the spatio-temporal activity within a spiking network
without the need to convert the activity to another representation that
inevitably loses much of the information.

\subsection{PNG Detection Algorithms}
Initial methods proposed for the detection of PNGs \cite{Izhikevich2006,Martinez2009} were predominantly
brute-force approaches that tested every combination of input stimulus
that was possible to trigger a PNG activation. This would be done in a two
stage process. Firstly, adapted PNGs would be determined by optimally
stimulating every set of three group triggering neurons and record the
resulting activity for each combination. Secondly, activated PNGs would
be detected in the spiking activity by pattern matching each triplet of
spikes against the stored adapted PNGs. The inefficiency of these
procedures prohibited their wide application. More recently, alternative 
methods have emerged that improve the
efficiency of detecting active PNGs \cite{Sun2015} or a probabilistic 
\emph{fingerprint} of polychronous activity \cite{Guise2014}.

\subsection{Motivation for a Polychronous State}
From their introduction, the general concepts of Polychronous activity and Polychronization
of neural networks has been distinct from the specific structure of a PNG \cite{Izhikevich2006}. The latter were
used to explore the structural nature of spatio temporal neural activity as well as a
conceptual link to Neural Darwinism and the Theory of Neural Groups Selection.
\\
It should be noted that in many respects, PNGs have an arbitrary definition and one that comes
with a few restricting limitations:
\begin{enumerate}
\item PNGs must be triggered by precisely three anchor neurons connected through a single root neuron.
\item The minimum network path length of a PNG must be seven or other arbitrary number.
\item Identification of a PNG must happen in a silent, noiseless network.
\item Network boundaries for a PNG are fuzzy, they must be truncated for reliable active detection.
\end{enumerate}
While PNGs have their role for structural and network capacity analysis, when the task is real-time
pattern recognition, the disadvantages outweigh the benefits of using them. For this use-case,
a method is needed for quantifying the polychronous state of a network at any point in time during the
presentation of a pattern, or at the end. In the section that follows, we introduce an algorithm
to form a polychronous encoding during the computation of neural spiking activity that can fill this
role.

\section{A Minimal Polychronous Model}
We simplify the requirements for a polychronous pattern in a number of ways. Firstly,
the atomic unit of polychronization is defined to be a single spike, rather than a
groups of neurons. Getting rid of the group structure also rids us of the arbitrary
boundary conditions that determine the neurons within the group, i.e. precisely three
triggering anchor neurons and a lower threshold on the maximum path length of the
PNG. Secondly, a polychronous pattern is solely based on neural activity, not on the
structure of the network or synaptic strengths. This removes the dependency of
searching for structural and dynamical PNGs before detecting activated ones.

\begin{figure}[ht]
\hfill
\begin{center}
\includegraphics[width=4in]{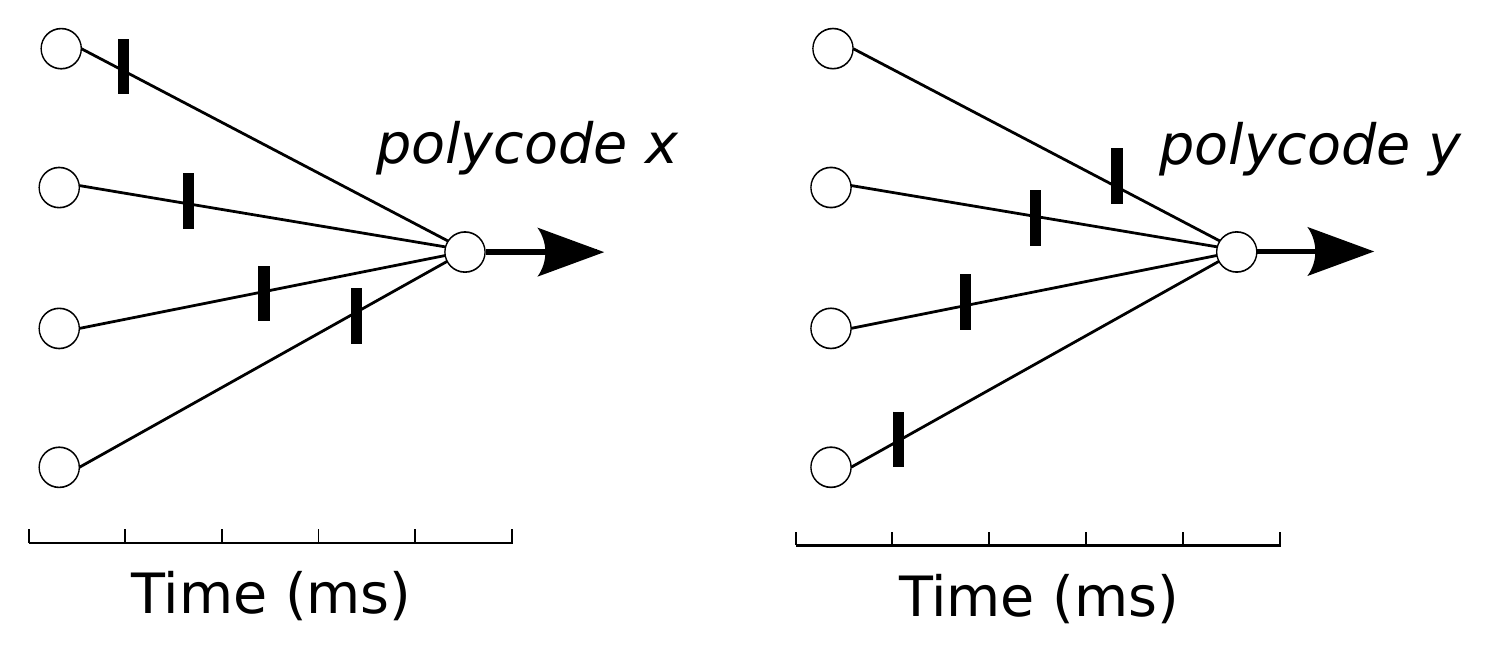}
\end{center}
\caption{Each polychronous pattern is given a unique code, a \textit{polycode}.
The \emph{polycode} is generated based on the precise ordering of pre-synaptic spikes
that cause a post-synaptic neuron to spike. The timing of each spike is not
used, just the relative order of pre-synaptic spike transmissions.}
\label{fig:polycode_illustration}
\end{figure}

We observe that during the activation of a PNG, the pre-synaptic sequence of spikes
will be fixed for each generated spike, otherwise it would constitute a different PNG.
Hence, we define a generated spike with a fixed order of pre-synaptic input spikes 
to be a distinct polychronous pattern. This is illustrated in Figure \ref{fig:polycode_illustration}. 
In our proposed scheme, each ordering of pre-synaptic spikes produces a different code
that is unique to a polychronous pattern. This process is described in the next
section on detection. As the generation of these codes is central to our method,
we refer to these minimal polychronous patterns as \textit{polycodes} in order to 
distinctly identify them from PNGs.

We can say for certain that a particular polycode will always activate when a 
particular PNG activates. Therefore, if PNG X is active whenever stimulus
X is presented, then polycode X will also be active. Of course, polycode X has the
potential to activate when PNG X does not. This logic means that
polychronous codes have the same response consistency properties of PNGs but 
that individual polycodes are not guaranteed to be as representationally 
selective. Thus, a polycode is a sub-component of a PNG.

\begin{figure}[ht]
\hfill
\begin{center}
\includegraphics[width=4.5in]{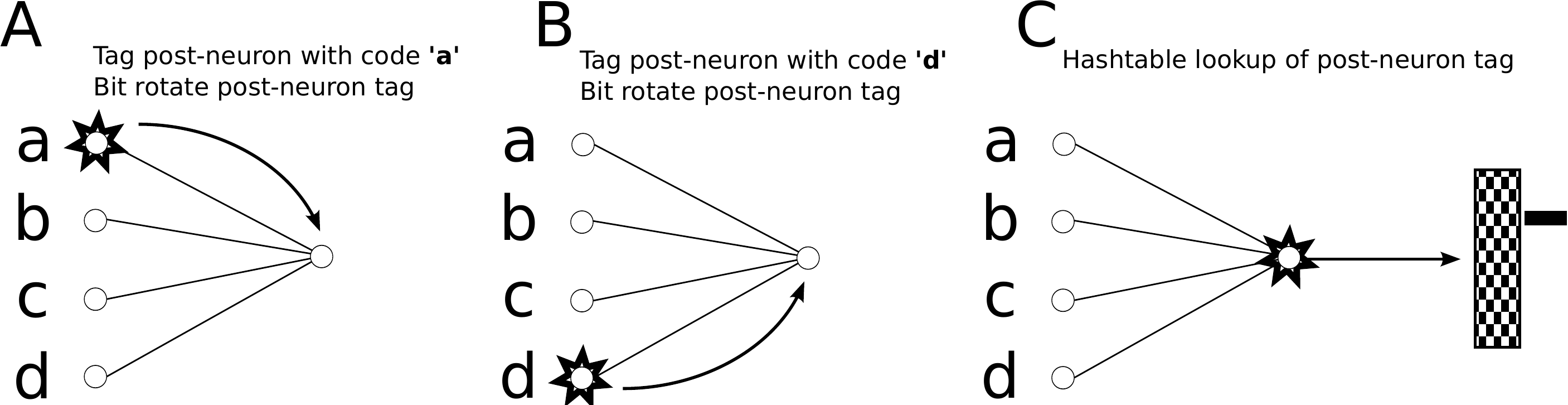}
\end{center}
\caption{Depiction of the method for PC detection.
{\bf A,B:} Pre-synaptic activity causes the post-neuron to be
\emph{tagged} with the pre-neuron codes. Between each tag the bits
are rotated, which means each order of tagging leads to a
different code.
{\bf C:} When the post-neuron spikes, the current tag code
is used as a hash key in a hashtable lookup. Each cell is
a unique temporal sequence.}
\label{fig:polycode_method}
\end{figure}

Figure \ref{fig:polycode_method} illustrates the formation of a polycode and
its subsequent activation when the post-synaptic neuron spikes. The algorithm
for the tagging and bit rotation parts are thoroughly explained in the methods
section.

The theoretical capacity of polycodes in a given network  is
$(N \cdot S!)$ where $N$ is the number of neurons and $S$ the
number of synapses per neuron. Due to the vast capacity for
any networks with more than about ten synapses per neuron, the
bit precision of the polycodes are the limiting factor. Depending
on whether a 32 bit or 64 bit code is selected, the capacity would
be about 4 billion or 18 billion billion, respectively. The chances
of polycode collisions within this space are determined by the hash
spread function and the number of polycodes that occur in a given
spiking network. In our experiments, explained in later sections,
there are about half a million polycodes observed which falls within
an acceptable range to avoid collisions with either bit precision.

\section{Methods}

\subsection{Neural Network}
The neural network model used in this work follows the implementation defined
in \cite{Izhikevich2006}. Recurrently connected neurons, denoted by $L$ are 
stimulated by the inputs directly as injected current, $I$, that perturbs the 
membrane potential modeled with a simple model \cite{Izhikevich2003}. 
This method for modeling the spiking activity of a neuron is shown to reproduce most
naturally occurring patterns of activity \cite{Izhikevich2004a}. The real-valued
inputs are normalized between $0$ and $1$, which are multiplied by a scaling
factor of $20$ before being injected as current into $L$. Input connections
project from a $16x16$ grid of pixels, each stimulating a single excitatory neuron. 
The network activity dynamics are then simulated for $30ms$.

For our experiments the network consists of 320 spiking neurons with the ratio 
of excitatory to inhibitory as 256:64. Neurons are pulse-coupled with static synapses i.e. the delta
impulse (step) function. Connectivity is formed by having $N^2 \cdot C$ synapses that 
each have source and target neurons drawn according to uniform random distribution, 
where $N$ is the number of neurons and $C$ is the probability of a connection 
between any two neurons.
Weights are drawn from two Gaussian distributions; $\mathcal{\mathcal{{N}}}(6,0.5)$ 
for excitatory and $\mathcal{{N}}(-5,0.5)$ for inhibitory. All parameters for excitatory 
and inhibitory neuron membranes are taken from \cite{Izhikevich2003}. The equations 
for the membrane model are as follows:

\begin{equation}
    v'=0.04v^{2}+5v+140-u+I
\end{equation}

\begin{equation}
    u'=a(bv-u)
\end{equation}

With the spike firing condition: 

\begin{equation}
    \text{if}\quad v>30mV\quad\text{then}\quad\begin{cases} v\leftarrow c\\ u\leftarrow u+d\end{cases}
\end{equation}

\subsection{Task and Stimuli}

\begin{figure}[ht]
\hfill
\begin{center}
\includegraphics[width=4.5in]{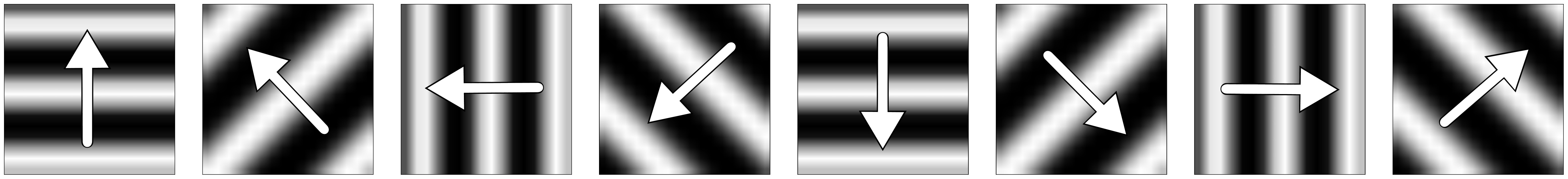}
\end{center}
\caption{Moving directional bars that are used as stimuli in a task that
tests the directional selectivity of simple cells. An example of the use
of this type of stimuli can be observed in neuroscience studies on low
level circuits in the visual cortex \cite{Mazurek2014}.}
\label{fig:stimuli}
\end{figure}

A simple visual task to determine direction selectivity of motion is taken from a
recent study in computational neuroscience \cite{Mazurek2014}. This task is suitable
for the small, cortical column sized \cite{Habenschuss2013} network that we are
working with that is connected directly to the visual stimuli -- i.e. low in the
cortical hierarchy. The inputs consist of moving bars that take one of eight directions, 
0$^{\circ}$, in 45$^{\circ}$ increments, through to 315$^{\circ}$. Static images of these 
input patterns are visualized in Figure \ref{fig:stimuli}. The frame dimensions 
are $16x16$ pixels, each one is used as an exclusive input to a single excitatory 
neuron. This direction selectivity task is used in later sections to establish the 
representational ability of polycodes and an example of their use in pattern recognition.

\subsection{Polychronous Pattern Detection}

Pseudo code that describes the algorithm to generate polychronous codes
is given as follows:

\begin{Verbatim}[commandchars=\\\{\},codes={\catcode`$=3\catcode`_=8}]
\textit{// to be called every simulation time-step}
\textbf{if} ($this$ neuron spikes) \{
	\textbf{for each} post-synaptic neuron $post$ \{
		XOR($post$.code, $this$.tag)
		ROTL($post$.code, 1)
	\}
	\textit{// ignore spikes caused by external input}
	\textbf{if} ($this$.code != $this$.tag) \{
		$hashmap$.emplace($this$.code, $label$)
		// reset code
		$this$.code = $this$.tag
	\}
\}
\textit{// only combine causal activity in the code}
\textbf{if} ($this$.Vm < 0) \{
	$this$.code = $this$.tag
\}
\end{Verbatim}

This code can be integrated with the core of a time-step based spiking
neural network simulation. The $tag$ values are randomly generated
binary strings that are fixed for each neuron in the network.
The $code$ values are the polychronous codes that are initialized per
neuron, as the $tag$ and subsequently updated according to the
pseudo code.

On the occurrence of a spike, two things are
triggered. Firstly, all of the codes at the post-synaptic neurons are
XOR'd with the pre-synaptic neurons tag code and their bits are rotated.
This is the step that generates evenly spread and likely unique valued codes
for each combination of pre-synaptic activity that causes a spike. 
Secondly, the polychronous code value for the neuron that has just spiked 
is used as a hash key in a lookup table. This should only occur if the code 
is different from its initial value, otherwise the spike will have just been 
caused by external input. Information about the pattern can be stored in the cell,
such as a class label or a repetition value. The last part of the 
pseudo code is run every time the neuron membrane activity is updated. It
resets the polychronous code to its initial value if the membrane potential
crosses a lower threshold so that the code only reflects pre-synaptic 
activity that had a causal role in generating a spike.

\section{Results}

\subsection{Stability of Repeating Patterns}

\begin{figure}[ht]
\hfill
\begin{center}
\includegraphics[width=4.5in]{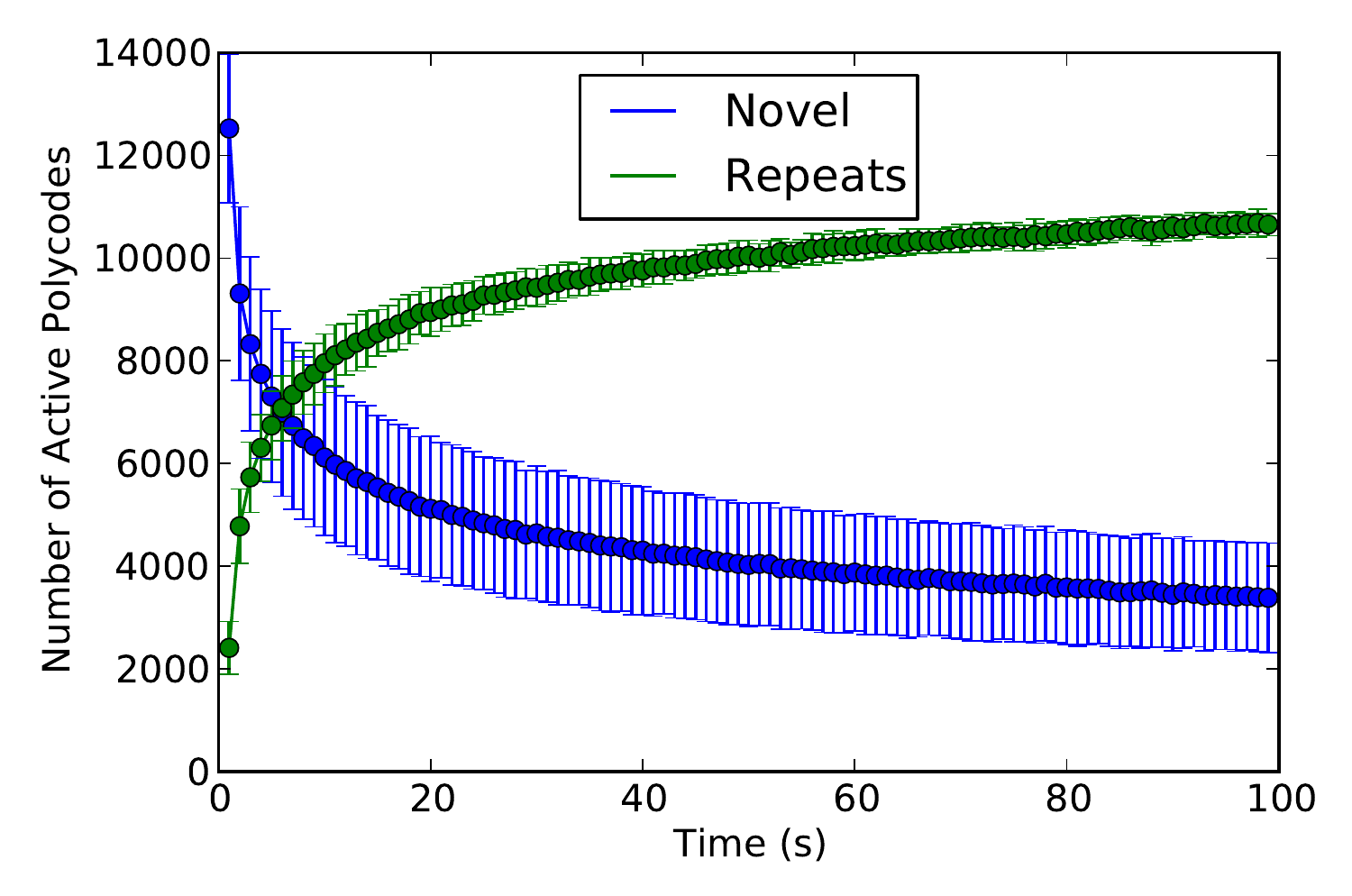}
\end{center}
\caption{Over the course of presenting the eight moving stimuli for 100
seconds, the number of novel and repeating polychronous patterns are
recorded. Bars indicate standard deviation over ten trials.}
\label{fig:stability}
\end{figure}

The repeatability of patterns are the fundamentally required property for
them to form representations of input stimuli \cite{Vertes2010,Guise2013}.
Initially, all patterns will be newly registering and it will take time for
repeats to occur. Figure \ref{fig:stability} plots the occurrence of novel
and repeating polycodes while a directional stimulus is presented over
100 seconds. Each point in the graph is an average of the eight input
stimuli.

In each second of simulation, there are about $15k$ polycodes active.
That corresponds to a neural network activity level of just under $5\%$
on each millisecond time-step. The number of repeating polycodes overtakes
the number of novel ones at the six second mark. Eventually, there are
over $10k$ repeating polycodes, which have the potential for representational
consistency. The remaining level of $3.5k$ novel polycodes indicates there
is continual source of new patterns in the neural activity, given that the input
patterns are uniformly repeating. This continual occurrence of new patterns must be
due to the repeating inputs convolving with the fading memory of the spiking
activity.

\subsection{Representational Selectivity}

\begin{figure}[ht]
\hfill
\begin{center}
\includegraphics[width=4.5in]{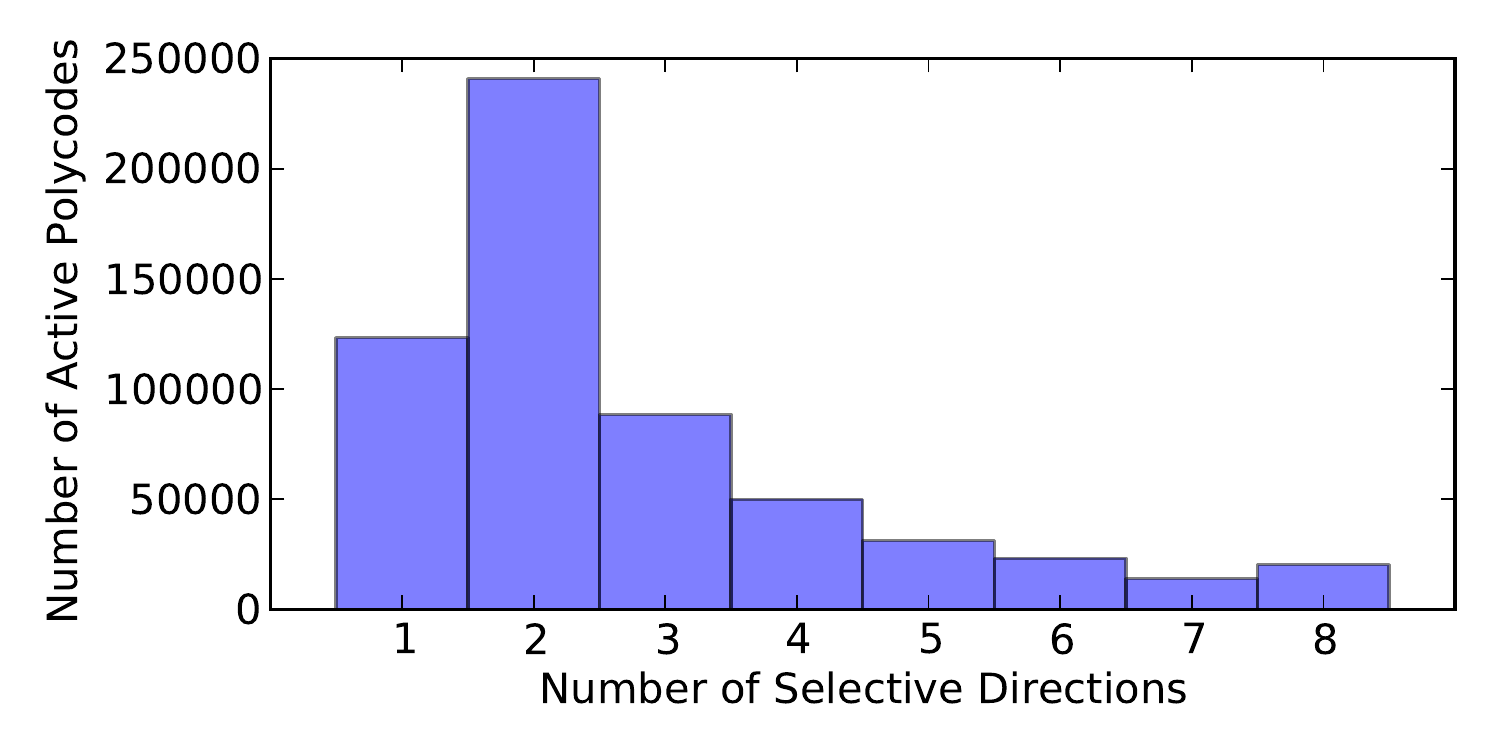}
\end{center}
\caption{Directional selectivity of all the polychronous patterns
detected within 100 seconds of simulation. The selectivity relates 
to how many directions a polycode activates in response to.}
\label{fig:selectivity}
\end{figure}

Activation consistency alone is not a sufficient condition for a representational
system. Polycodes also must be shown to be selective, i.e. are only active when
a subset of the input types are presented, ideally a single type of input sample.
Figure \ref{fig:selectivity} plots the number of polycodes that are active for each
quantity of input sample direction.

The bulk of the polycodes are active when two directions of input are presented
as stimuli. However, there are a significant number, above $100k$ in total, that
are only active for a single particular direction. Also, there are comparatively
few polycodes that are active for any direction which indicates that the coding
method is highly sensitive to the input stimuli.

\subsection{Pattern Recognition}

The previous properties of polycode occurrence, consistency and selectivity are
now utilized in a pattern recognizer.

\begin{figure}[ht]
\hfill
\begin{center}
\includegraphics[width=4in]{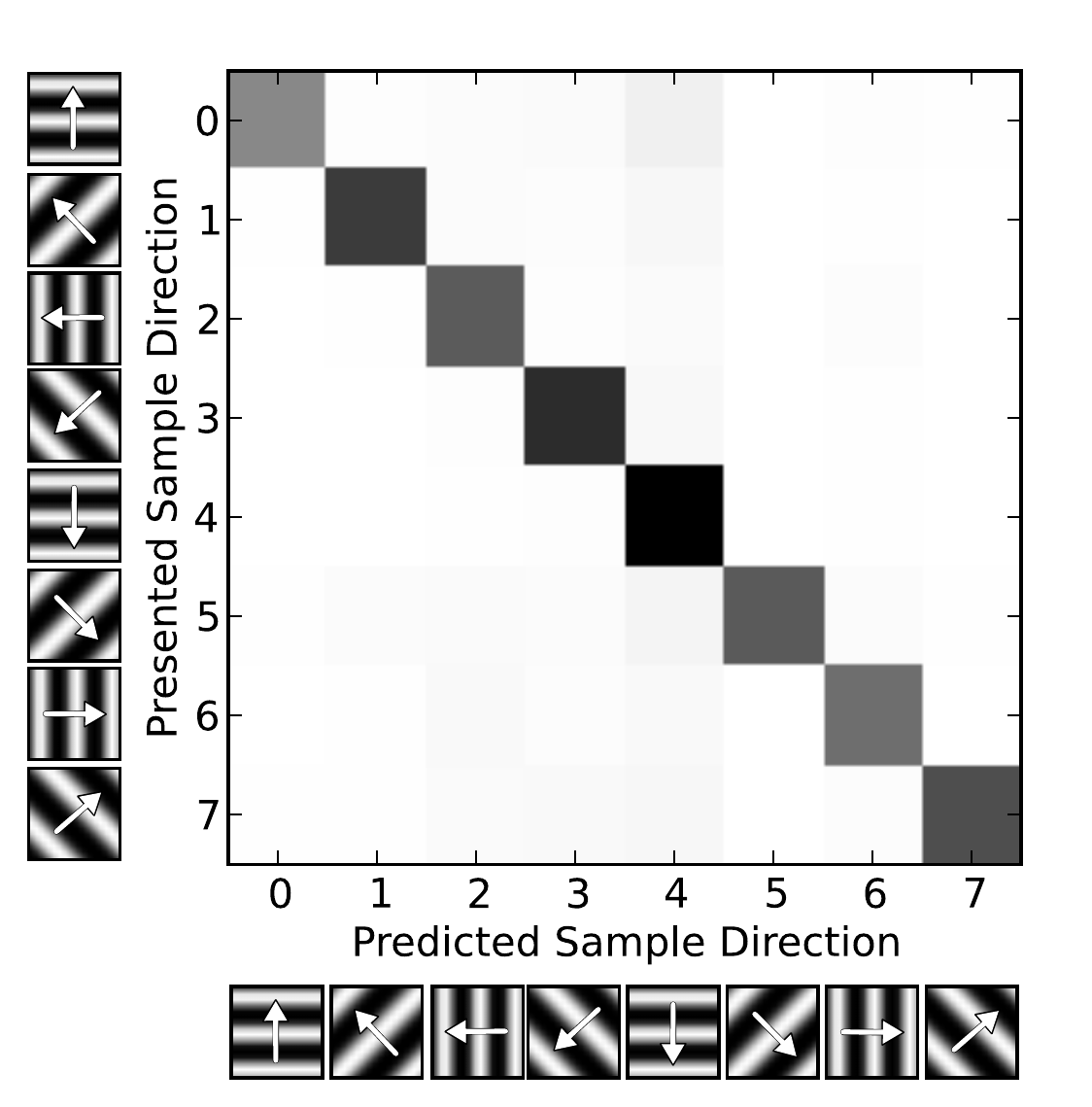}
\end{center}
\caption{Prediction vectors formed based on the repetition of polycodes in response to
each direction of moving bars. Input samples are indicated along the y-axis and
the predicted response based on the activated polycodes is indicated for each
direction along the x-axis.}
\label{fig:prediction}
\end{figure}

During the training phase, 100 second sample of
each directional input is presented. Whenever a polycode is active, two values are
stored at the corresponding hash table cell: ${directionLabel, repeats}$. Upon
a hash table lookup $repeats$ is incremented if $directionLabel$ matches
and is decremented otherwise. If $repeats$ goes down to zero, the $directionLabel$
switches to the current sample's direction. In the classification phase, an $nDirection$ 
dimension prediction vector, $pred$, is formed in which 
$pred[directionLabel]=\sum log_{2}(repeats)$. Finally, the predicted direction
is determined by $max(pred(\cdot))$. Figure \ref{fig:prediction} plots the prediction vector for each of the presented samples (along the y-axis) with the $log_{2}(repeats)$ values for each $directionLabel$ (along the
x-axis).

This simple pattern recognition method manages to amplify the effect of the polycodes that repeat in
response to particular patterns and thus forms the basis of an effective classifier for this
low-level visual task.

\subsection{Efficiency}

The detection of minimal polychronous patterns as proposed in this article imposes an overhead
throughout the spiking simulation, instead of running as an off-line process that scans through
spike data generated by the simulation. Whenever a spike occurs, a few extra instructions must
execute per synapse along with a single hash table lookup.

\begin{figure}[ht]
\hfill
\begin{center}
\includegraphics[width=4in]{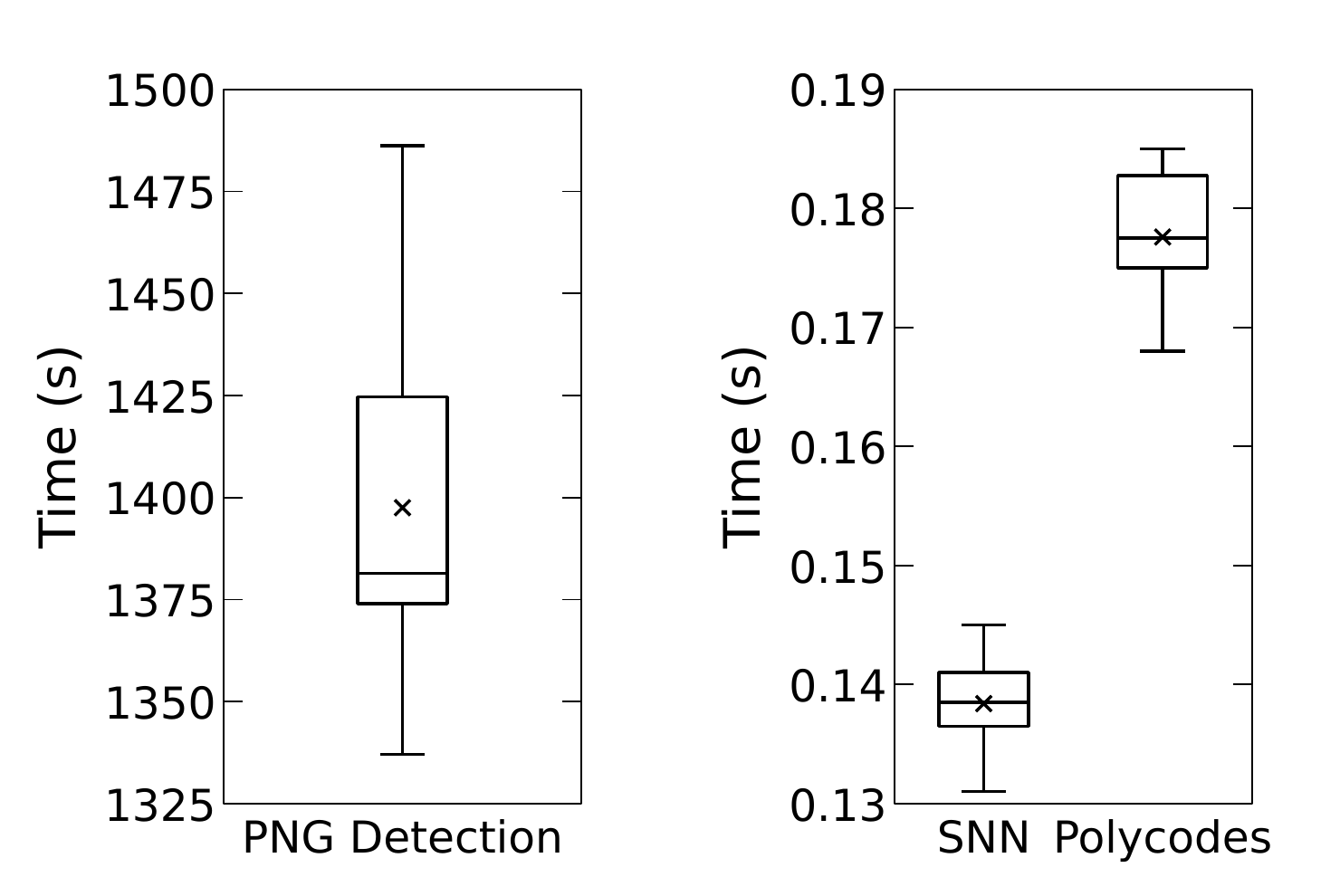}
\end{center}
\caption{A benchmark of computational efficiency between PNG and polycode detection. Each box
represents ten simulations with the random seed set by the clock. Left: The time taken for
a single pass of PNG detection. Right: Time taken to simulate ten seconds of spiking activity
plotted with the same length spiking simulation including polycode detection.}
\label{fig:benchmark}
\end{figure}

The efficiency of the proposed algorithm cannot be directly compared to traditional PNG detection
because the patterns detected by it are substantially simplified when compared to the structural
and temporal information contained implicitly within a PNG. However, for a reference, it takes
about 23 minutes to perform one pass of PNG detection using the code distributed along with the 
introductory paper of polychronization \cite{Izhikevich2006}. This stands in contrast to the
39ms overhead per ten simulated seconds imposed by our minimal polychronous pattern detection.
A comparison of the runtime efficiency between PNG and polycode detection is shown in Figure
\ref{fig:benchmark}. The overhead of polycode detection can be seen in the right hand plot and
is four orders of magnitude smaller than PNG detection time shown on the left.

\section{Discussion}

\subsection{Advantages}
The detection of polychronous codes provides a rapid way to detect precise spike-timing
patterns. Previously, there was a choice: inefficiently detect PNGs \cite{Martinez2009},
decode spike sequences into real-values \cite{Paugam2012}, or perform some computation of distance between
the spike sequences themselves \cite{VanRossum2001}. The latter two options do not have the ability to
reliably distinguish spatio-temporal spiking patterns from their output values.
This minimal method of polychronous code detection is even more efficient than recent alternative
forms of PNG detection \cite{Guise2014,Sun2015}, which are themselves vast improvements over the initial
algorithms. The fastest of these alternative methods \cite{Guise2014} requires several hundred extra seconds of spiking simulation per stimuli in order to detect the equivalent of a polychronous response.

We have shown through a simple visual motion sensitivity task that polycodes have the properties
of response consistency and selectivity as required by representational systems. These properties
have been exploited in the construction of a pattern recognizer that works on polycodes
directly. The minimal model of polychronization described here also has the advantage of removing
many arbitrary constraints on the definition of a PNG. A polychronous pattern need no
longer require triggering by precisely three anchor neurons. Also, the arbitrary threshold
imposed by a minimum longest network path length is not present.

\subsection{Limitations}
The minimal model of polychronization proposed has none of the structural information that
PNGs contain implicitly. This is a particularly serious limitation if the intent is to
analyse structural properties of a network through detected PNGs. However, the works using
polychronization to date have largely used PNGs as a representational model that only relies
upon their formation and occurrence in response to input stimuli \cite{Paugam2008,Guise2015thesis}, 
not their structural properties.

Another limitation is the theoretically reduced selectivity of polycodes as compared with PNGs.
By definition, the polycodes have the same or better response consistency of PNGs but this
does not hold with selectivity. In fact, it is very likely that polycodes are far less
selective than PNGs due to their far simpler activation requirements. This problem would need
to be mitigated by building a representational system around populations of polycodes instead
of a paradigm of $1 \ class = 1 \ code$.

\subsection{Future Work}
The model and methods outlined in this work is just the basis of a simpler,
more efficient form of polychronization. There are a number of key areas that
are in need of investigation using this new methodology.

\par

\begin{description}
 \item[Plasticity forming representations.] Our minimal model of polychronization can
 be applied to any spiking network activity, unlike the original model, which
 relied upon the evolution of synaptic weights through STDP \cite{Izhikevich2006}. However, 
 plasticity has a central role in the functional self-organization of the nervous
 system in response to environmental stimuli. Therefore, it is essential to
 investigate the emergence of polychronous codes in response to specific input
 patterns while the synapses are adapting according to plasticity. In particular,
 we stress the importance of analysing the representational properties of these codes
 to determine if unsupervised synaptic adaptation can improve their response consistency
 and selectivity.
 
 \item[Hierarchical polychronous patterns.] The experiments presented here use a single
 recurrently connected network to obtain a polychronous response from the input stimuli.
 This is analogous to a single cortical mini-column \cite{Habenschuss2013} that might be detecting one type
 of pattern in the mammalian brain. For a truly powerful representational system, it is
 expected that pattern recognizers work in a massive hierarchy in which higher levels
 respond to increasingly abstract features of the input \cite{Kurzweil2012}. In terms of experimentation,
 the response consistency and selectivity of polychronous codes could be measured for
 a series of connected layers of networks which each use the previous networks output as
 its own input. It would be expected that consistency and selectivity increases with 
 additional layers. This would indicate a higher degree of invariance as well as the ability
 to recognize higher level patterns, more general than localized spatio-temporal patterns.
 
 
 \item[Regression using polychronization.] The representative nature of PNGs and polychronous
 codes make them particularly suitable for classification tasks. Regression problems generally
 require a quantitative output that can be combined with trainable real-valued parameters in
 order to approximate a desired signal. We take inspiration from the Cerebellar Model 
 Articulation Controller (CMAC) from autonomous robotics \cite{Albus1975}.
 This model is arguably not a network at all, but rather combines sensor input 
 with internal state and maps the result to a set of cells through a hash table. The CMAC model 
 enables regression by storing a real-value at each cell of its hash table. Activated 
 cell values are trained by
 iterative gradient descent. We propose that it is possible to use this regressive model
 when the hash table cells are determined with polychronous codes, thus enabling function
 approximation in addition to pattern recognition that is typically associated with 
 polychronization.
\end{description}

\par

It is hoped that the simple approach and algorithm presented in this paper can facilitate
investigations to the above areas as well as others that the authors cannot foresee.

\section*{Acknowledgments}
This work was supported in part by an Engineering and Physical Sciences Research Council's Doctoral Training Grant through University of Surrey.
\section*{References}
\bibliography{refs}

\end{document}